# Complexity of Representations in Deep Learning

Tin Kam Ho
IBM Watson Health
Yorktown Heights, NY, USA
tho@us.ibm.com

*Abstract*—Deep neural networks use multiple layers of functions to map an object represented by an input vector progressively to different representations, and with sufficient training, eventually to a single score for each class that is the output of the final decision function. Ideally, in this output space, the objects of different classes achieve maximum separation. Motivated by the need to better understand the inner working of a deep neural network, we analyze the effectiveness of the learned representations in separating the classes from a data complexity perspective. Using a simple complexity measure, a popular benchmarking task, and a well-known architecture design, we show how the data complexity evolves through the network, how it changes during training, and how it is impacted by the network design and the availability of training samples. We discuss the implications of the observations and the potentials for further studies.

## I. Introduction

Deep learning methods are believed to be able to learn new representations of the input that can capture the essential data characteristics important for a target task like classification. When a trained network achieves a satisfactory level of accuracy, it is believed that the representations learned in the training process must also be good. With some tasks like image recognition, the intermediate representations produced by certain layers can be visualized and inspected for assurance that they do in fact capture essential features of the input images.

There are issues with these ways of justifying or confirming the learning achieved in the process. Consider a classical example task of handwritten digit recognition. Visual inspection of the intermediate image features assures that the learned representations agree with human intuition of what is important in the images, or how unimportant variations are masked out, but it does not give a metric that can be used to determine how such an agreement with intuition helps with classification. Assessment of end-to-end classification accuracy affirms the overall effectiveness of the trained network, but does not give much insight about the effectiveness of learning in the internal layers.

One way to analyze the characteristics of classification data is through data complexity measures [1][2]. This family of methods attempts to relate the intrinsic geometric and topologic properties of a dataset to the difficulty of a classification problem defined on the dataset. If a deep neural network is to achieve good classification accuracy, the input must be mapped by the network to a space where the classification task becomes easy (e.g. doable by taking simple maximum of a decision function's value). In many cases the mapping goes from a very high-dimensional input space to a very low dimensional output space. The network needs to address the two issues at the same time: compress the objects to a lower dimensional space and also push the objects of different classes to be maximally separated. These two objectives are fulfilled by a cascade of multiple stages of mappings carried out in the internal layers. To better understand the inner working of a deep network, it is useful to examine in detail how each layer contributes to these objectives. Given a specific network design, the dimensionality changes through the layers are fixed. The effect of network training is left to be on promoting class separability. To follow how this happens, we can analyze the contribution of each layer to the change in class separability using data complexity measures. By monitoring the changes in complexity as the data pass through the network layers, we may gain more detailed knowledge on the role played by each layer as training progresses. This can serve as a start to open up the "black-box" that is often an impression given by deep neural networks.

## II. Classification Data Complexity

Given a representation of a set of objects in a feature space, certain "natural" groupings of individual objects can often be observed using a metric of proximity. Assignment of class labels according to a certain task goal adds another attribute to the objects. The class assignment may or may not align well with the pre-existing natural associations. The classification task is easy when there is good alignment – for example, when the class labels match well with membership of clusters found through unsupervised learning with a "natural" metric. Notice that here we use the term "natural" to refer to those metrics that are commonly used (e.g. Euclidean distance, cosine distance) and do not require special tuning or complex transformations in their computation. In the opposite case, if class labels are assigned in a way not predictable by the natural associations, the classification task is difficult. This can occur if the classes are determined by some artificial or random processes, or by reliance on attributes of the objects not represented in the given features.

Observing that classes are easy to learn when they align well with naturally occurring clusters, one may consider using simple processes and metrics often used in unsupervised learning to characterize data complexity. In [3] it is shown that a complexity measure that is the strongest predictor of classification performance is the nearest neighbor error of a dataset, which stands out among 22 measures that participate in a regression analysis of the correlation of classification accuracy with these measures. In this work we start with this basic complexity measure. We use it to analyze the intermediate representations obtained in deep learning, for the well-known task of handwritten digit recognition with the MNIST dataset. It is our belief that similar analysis can be carried out with other measures (e.g. [4]) and other learning tasks. Our expectation is that in-depth analysis of the internal representations achieved by each layer of feature mapping in a deep network can shed some light on the role that the layer plays in the network.

## III. DEEP LEARNING EXPERIMENT SETUP

We start with the well-known task of handwritten digit recognition with the MNIST dataset (formatted as 32x32 pixels). One of the earliest studied architectures for this task is the leNet5 [5]. Here we adopt an implementation in PyTorch as published in [6][7]. Specifically, the network is a sequential concatenation of the layers as shown in Figure 1. A softmax function is applied to the final layer output; an input image is assigned to the class with a maximum value from this function. We adopt the same parameters as in [7] (random seed = 42, learning rate = 0.001, training batch size = 32, Adam optimizer, cross entropy loss). In this implementation, layers "conv1" through "tanh3" are said to serve as feature extractors, and layers "linr1" through "linr2" are said to be the classifier. We refer to the layer denotation in Figure 1 in the subsequent discussions.

To examine the intermediate representations of the data, we use the "register_forward_hook" function in PyTorch to save the input and output tensors at each network layer. After the network is trained to a desired number of epochs, we activate the function hook and pass each image in the dataset

```
self.conv1 = nn.Conv2d (in_channels=1, out_channels=6,
                       kernel_size=5, stride=1)
self.tanh1 = nn.Tanh()
self.pool1 = nn.AvgPool2d (kernel_size=2)
self.conv2 = nn.Conv2d (in_channels=6, out_channels=16,
                       kernel_size=5, stride=1)
self.tanh2 = nn.Tanh()
self.pool2 = nn.AvgPool2d (kernel_size=2)
self.conv3 = nn.Conv2d (in_channels=16, out_channels=120,
                       kernel_size=5, stride=1)
self.tanh3 = nn.Tanh()
self.linr1 = nn.Linear (in_features=120, out_features=84)
self.tanh4 = nn.Tanh()
self.linr2 = nn.Linear(in_features=84, out_features=10)
```

Fig. 1. Implementation of leNet5 in PyTorch.

through the network in evaluation mode. The resulting tensors containing the intermediate representations, together with the label of the input image, are saved for subsequent complexity evaluation. As discussed before, in this study the chosen complexity measure is the nearest neighbor error rate (in Euclidean metric), which is estimated for each dataset in a relevant representation using the leave-one-out method.

## IV. COMPLEXITY OF LEARNED REPRESENTATIONS

We first study the network trained with the full set of MNIST training data (60000 images). Figure 2 shows the plot of complexity of the training (a) and testing (b) datasets at the entry and exit of each network layer. For each of the 15 training epochs, the change in complexity over the network is shown as a colored line. On the far right we also show the end-to-end network's classification error on the same dataset.

We can observe the followings:

- The chosen measure of data complexity correlates well with the end-to-end network's classification error, and the complexity drops together with classification error as training progresses over the epochs, tuning the decision boundaries of the network to the class boundaries.

- The dataset's complexity generally decreases as the data

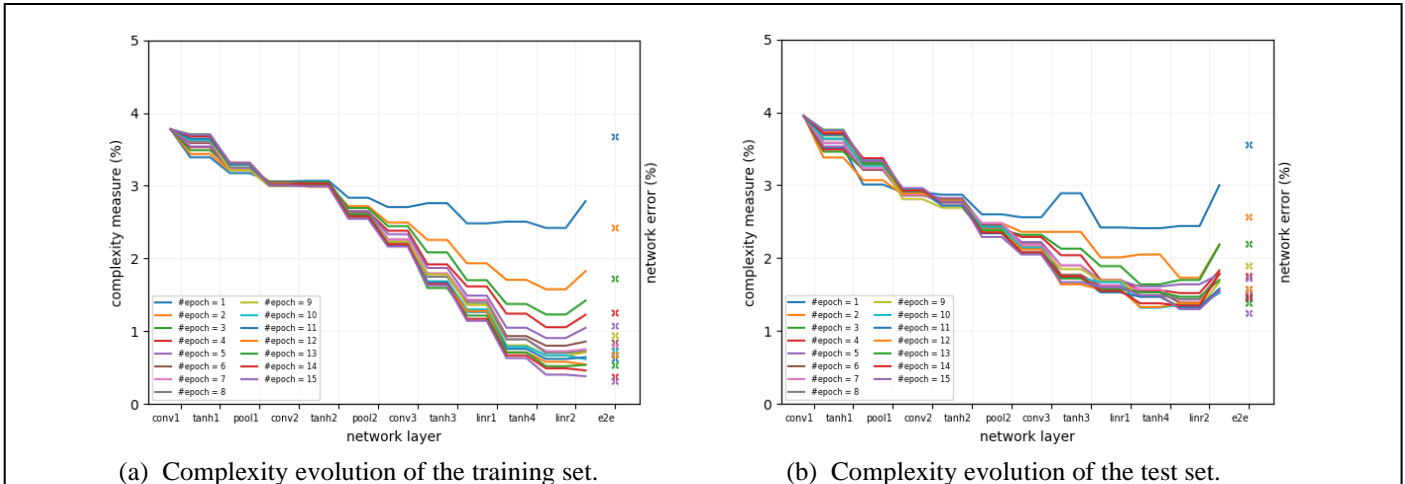

(a) Complexity evolution of the training set.    (b) Complexity evolution of the test set.

Fig. 2. Complexity of (a) training and (b) testing data as they enter and exit each layer of the trained network. On the far right, the end-to-end network error for the same dataset is shown for each epoch. To minimize the effect arising from the different sizes of training and testing sets, we estimate the training set complexity by dividing the data into 6 subsets each of the same size of the test set and then taking the mean of the 6 complexity values. The same process is followed for the other training set evaluations described in this paper.

go through each mapping defined by a network layer. There is no obvious distinction in this trend between the feature extraction layers and the classifier layers.

- Early epochs in training lead to some decreases in data complexity across most layers but far more substantially at the final layers. This could be an interaction of the nature of the mapping function (convolution, pooling, or linear, some with few or no adjustable parameters) together with its position in the network, if the amount of error back propagated to the initial layers diminishes on its way back.

- Complexity of representations for the training set drops faster that for the testing set. After the first few training epochs, the continuing drop in complexity of the training set is not seen in the testing set. In other words, beyond a certain training stage, the remaining effect of training is largely just fine-tuning the network to the training data, with little effect that can be carried over to the testing data. This overfitting behavior could impact more the tasks with sparse training data, which we see in the next section.

In this network, the initial layers like convolution and pooling are to capture the property of image data, where shapes are defined on combinations of pixel values over regular, pre-defined spatial patterns. The design is influenced by our understanding of images. This seems to go against the goal of using a deep network to learn the representation on its own. In the experiment these special layers appear to behave somewhat differently from the other layers in terms of complexity evolution. The complexity of the data mapped through these layers remains very stable over the training epochs.

## V. LEARNING WITH REDUCED TRAINING DATA

The MINST data contain a dense training set with 60000 images in 10 classes, capturing a rich set of variability in the digit shapes at a resolution of 1K pixels. Such dense sampling is not common with many other image recognition tasks with more complex object shapes, larger images, and/or many more classes. To better understand the effect of training sample sparsity, we reduce the training set to 1/6 of its size, i.e., to 10000 images, using a simple scheme of taking every $6^{th}$ image in the sequential order of the dataset, starting with the first image. We then retrain the network from scratch using this smaller training set. The intermediate representations learned through training are then analyzed using the same complexity measure.

Figure 3 shows the effect of the reduction in training data. Figure 3(a) suggests that the reduced training set is even harder than the test set, with a higher complexity at the entry to the first layer of the network. In early training epochs, the network incurs large end-to-end error (13% for training set and 11.2% for test set), though this is quickly improved as training progresses. After 15 epochs, the end-to-end error rate and the data complexity at the final layers, is at about 0.3% for the training set. However, this success does not carry to the test set, which has and end-to-end error rate staying above 2% (Figure 3(b)). From the complexity measures, we can also see that the adjustments in the per-layer feature mapping due to later epochs of training almost have no effect in improving the separation of the test data – the complexity values stay above 2%, whereas in Figure 2(b), we can see the complexity values continue to drop to below 1.5% in the final layers. This is an effect from the network overfitting to the smaller training data.

## VI. EFFECTS OF DROPOUT LAYERS

One way to combat overfitting is to introduce "dropout" layers [8], which tend to make the network's behavior more generalizable to unseen data, as reported in many empirical studies. Here we modify the network architecture by introducing 4 dropout layers as shown in Figure 4. A drop out rate of p=0.2 (p being the probability of an element to be zeroed) is employed at each dropout layer. As before, we train the network with either the full training set or the reduced training set, and then study the differences in the network behavior. Note that even though the general effect of dropout

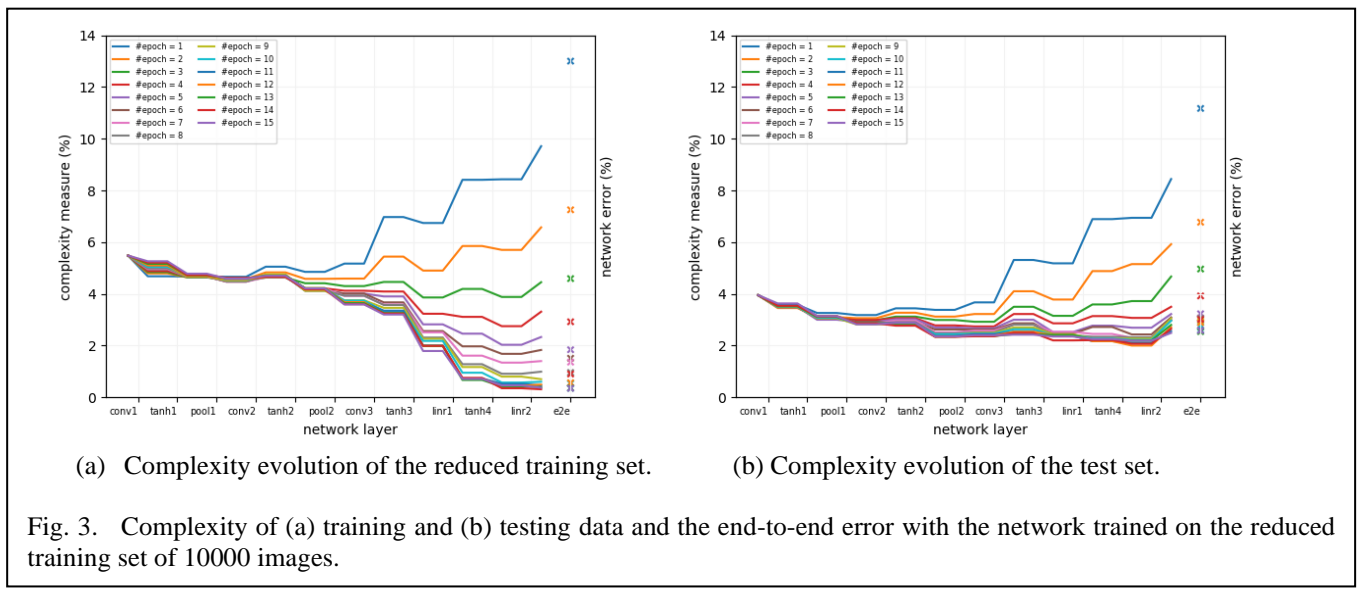

(a) Complexity evolution of the reduced training set.  (b) Complexity evolution of the test set.

Fig. 3. Complexity of (a) training and (b) testing data and the end-to-end error with the network trained on the reduced training set of 10000 images.

is well established, the details of how the dropout layers should be introduced are still being explored by many. For image recognition tasks, dropouts are commonly introduced after the convolution layers, and we follow such a practice.

Figures 5 and 6 show the behavior of the dropout-augmented network in terms of data complexity evolution over the internal layers and training epochs, when training uses the full set of 60000 images (Fig. 5) and the reduced set of 10000 images (Fig. 6), respectively. Several effects of dropout are apparent:

- The dropout-augmented network behaves more similarly on the training set and the test set, as Table 1 summarizes.

- It takes longer for the network to converge to the same level of accuracy, compared to the network without the dropout layers. While slower convergence means that

```
self.conv1 = nn.Conv2d (in_channels=1,out_channels=6,kernel_size=5,
                       stride=1)
self.drop1 = nn.Dropout2d (p=0.2)
self.tanh1 = nn.Tanh ()
self.pool1 = nn.AvgPool2d (kernel_size=2)
self.conv2 = nn.Conv2d (in_channels=6,out_channels=16,kernel_size=5,
                       stride=1)
self.drop2 = nn.Dropout2d (p=0.2)
self.tanh2 = nn.Tanh ()
self.pool2 = nn.AvgPool2d (kernel_size=2)
self.conv3 = nn.Conv2dv (in_channels=16,out_channels=120,
                        kernel_size=5, stride=1)
self.drop3 = nn.Dropout2d (p=0.2)
self.tanh3 = nn.Tanh ()
self.linr1 = nn.Linear (in_features=120,  out_features=84)
self.drop4 = nn.Dropout (p=0.2)
self.tanh4 = nn.Tanh ()
self.linr2 = nn.Linear(in_features=84, out_features=10)
```

Fig. 4. Modified architecture with 4 added dropout layers.

more computation resources are required for training, the improved robustness of the network could be critical in applications where training data are expensive to get, and failing could incur high costs, especially on unseen data that are just minor perturbations of known samples.

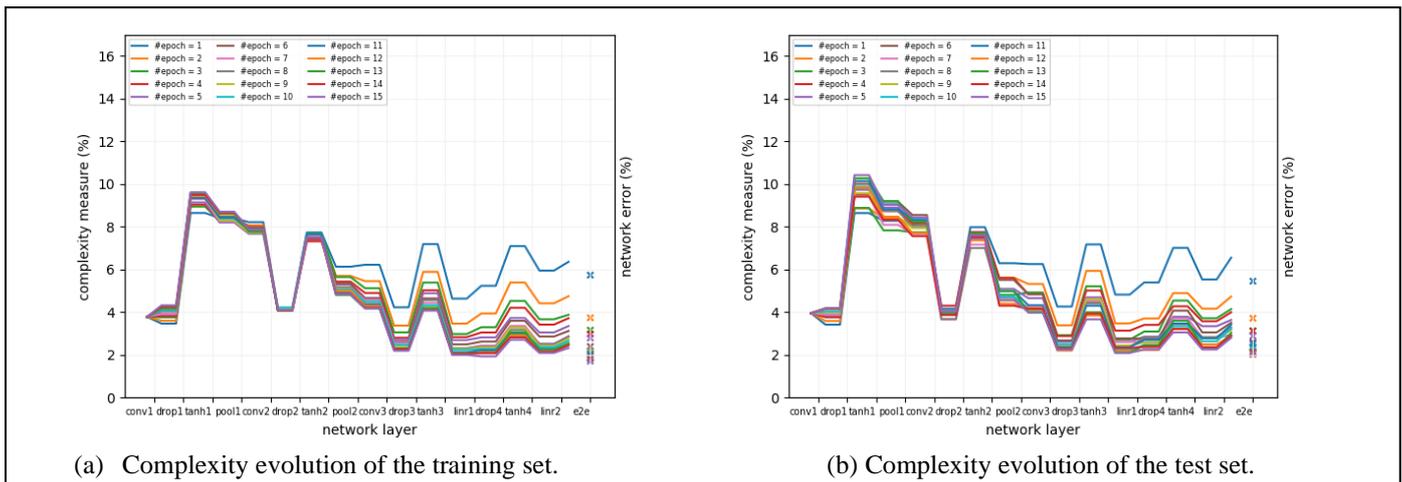

(a) Complexity evolution of the training set.    (b) Complexity evolution of the test set.

Fig. 5. Complexity of (a) training and (b) testing data and the end-to-end error with the drop-out enhanced network trained on the full set of 60000 images. The effects of dropout are seen as sharp rises in complexity across each of the dropout layers.

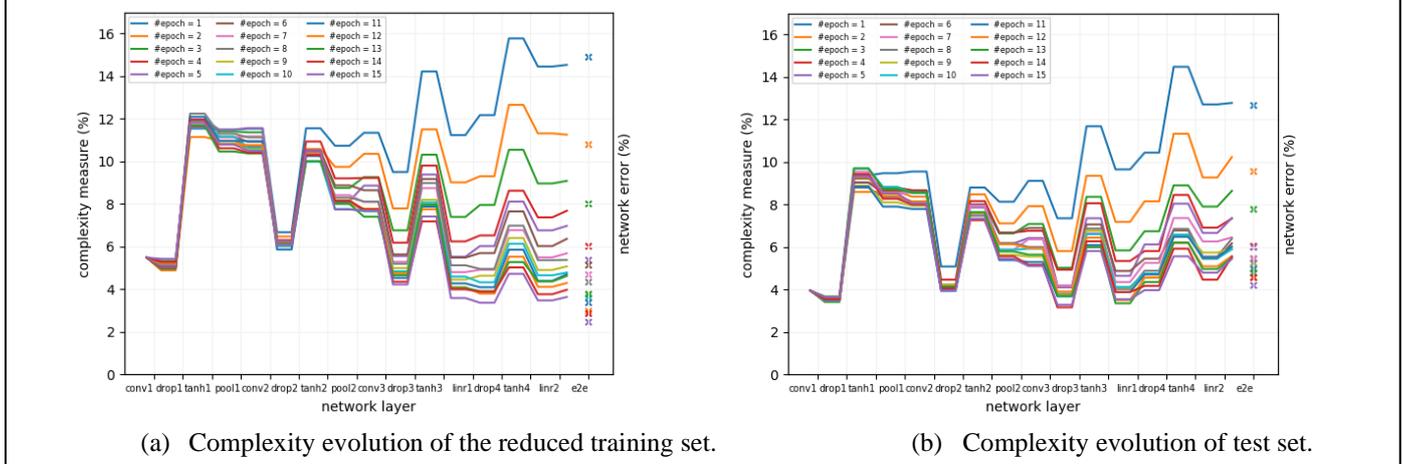

(a) Complexity evolution of the reduced training set.    (b) Complexity evolution of test set.

Fig. 6. Complexity of (a) training and (b) testing data and the end-to-end error with the drop-out enhanced network trained on the reduced training set of 10000 images. As in Figure 5, complexity rises sharply across each dropout layer.

- It is intriguing that the added dropout layers caused drastic increases in data complexity that the network needs to work hard to repair. This can be seen by comparing Figure 5 to Figure 2. Sharp rises and falls are seen in Figure 5, whereas in Figure 2 the complexity descent across the initial layers is largely monotonic. Similar differences can be seen in Figure 6 and Figure 3. The fluctuations in complexity, though tamed in magnitude, do not go away as training progresses to more epochs.

The trends in complexity evolution we see in this study resulted from the interactions of intrinsic data characteristics with many design choices, including the number of layers, the type of mapping used in each layer and the associated parameters (such as the sizes of convolution kernels), the optimization criterion, the learning rate, sizes of the learning batches, and others. While the trends match well with the prevailing impression and other empirical evidence associated with these design choices, it remains to be seen how sensitive the complexity behavior is to other variations in design.

## VII. DEEP NETWORKS AND MORE CHALLENGING TASKS

Can the complexity analysis be extended to deeper networks and more demanding problems? In this section we describe an application to another family of network architecture, ResNet [9]. We consider two designs of different depths as offered in [10]: ResNet-20 and ResNet-56 (see Appendices A and B in the Supplementary Materials for their detailed structures and parameters). Notably, these networks include many "batch normalization"[11] layers that are known to help reduce overfitting. We use the pre-trained network weights from [10] for the CIFAR-100 dataset [12]. The dataset includes 50000 training and 10000 test images labeled with 100 distinct classes, and is far more challenging than the MNIST data, as the object shapes captured are more variable, but fewer training samples are there for each class.

We pass the training or test data through the pre-trained network, capture the representation computed at each layer, and measure the complexity using the leave-one-out nearest neighbor error rate as before. Figures 7 shows the complexity evolution of both datasets for Resnet-20 and Resnet-56 respectively. We see some similar behavior of the network layers to those in leNet5: (1) contribution to reduction in data complexity is distributed over many network layers, but is concentrated on the final layers, (2) many of the "regularization" layers (batch normalization) cause rises in data complexity, and (3) the complexity differences between training and testing data are most significant in the final layers. In addition, the block structures in ResNet also produce many localized complexity evolution patterns at intermediate scales. These can be interesting subjects for further analysis.

## VIII. DISCUSSION

The observations give some insight on how the promise of representation learning is fulfilled in deep networks over different layers. The representations learned in a deep network is driven by the training objective that promotes class separation. Analyzing them in terms of their contribution to enhancing class separation (i.e., reducing data complexity), we can see how the effects of training and other parameter choices reach back to the early layers. This can be taken into account if the intermediate features are used to serve other purposes. For example, in natural language processing, lower dimensional representations learned in a first task is often "transferred" to serve another task, with or without fine tuning. It is our expectation that detailed analysis of the characteristics of the intermediate representations, and how they can be influenced by various training objectives and processes, can provide guidance to better control or adapt such representations to serve other goals.

The mappings implemented in deep networks serve to bring an object represented in a high-dimensional feature space to a decision function value in much lower dimensional space. In the MNIST digit example the ratio of dimensionality reduction is 1024 to 10. In this analysis we have seen that the contribution of each layer to this goal is not uniform. To a network designer, it could be interesting to know if there exists a type of "bottleneck" layer that prevents further improvements up or down stream, or a redundant mapping that does not contribute to increases in class separation.

The end-to-end network realizes a complex, composite mapping which, even if locally continuous, may connect or disconnect objects falling in different fragments of the input domain. With sparse training data, the known, labeled objects may be distributed far apart, leaving large unfilled space in between. In what way is it guaranteed that the network mapping will send a future object that happens to fall in the unfilled space to the desirable decision region? Or is that never guaranteed, as revealed by the discovery of many "adversarial" examples [13]? As deep learning makes its way into more critical applications, ignoring these issues could mean unknown and undesirable risks. We believe that a detailed understanding of these issues requires careful studies of the analytical properties of the intermediate mapping functions and the computable / quantifiable characteristics of the intermediate representations. In this work we have pursued a very small step of the later.

TABLE I. SUMMARY OF NETWORK END-TO-END ERROR RATE (%)

| % error on Train (TR) or Test (TE) Set | Basic leNet5 | | leNet5 + DropOut | |
|---|---|---|---|---|
| | *Trained with Full Set* | *Trained with Reduced Set* | *Trained with Full Set* | *Trained with Reduced Set* |
| epoch | TR  TE  TE-TR | TR  TE  TE-TR | TR  TE  TE-TR | TR  TE  TE-TR |
| 1 | 3.7  3.5  -0.2 | 13.0  11.2  -1.8 | 5.7  5.5  -0.2 | 14.9  12.7  -2.2 |
| 2 | 2.4  2.6  0.2 | 7.2  6.8  -0.4 | 3.7  3.7  0.0 | 10.4  9.6  -0.8 |
| 3 | 1.7  2.2  0.5 | 4.6  5.0  0.4 | 3.2  3.1  -0.1 | 8.1  7.8  -0.3 |
| 5 | 1.1  1.7  0.6 | 1.8  3.2  1.4 | 2.8  2.9  0.1 | 5.9  6.0  0.1 |
| 10 | 0.7  1.5  0.8 | 0.5  2.7  2.2 | 2.0  2.4  0.4 | 3.7  4.8  1.1 |
| 15 | 0.3  1.2  0.9 | 0.3  2.6  2.3 | 1.7  2.0  0.3 | 2.4  4.2  1.8 |

Just as an object can be represented by many different ways chosen by an observer or learned in a deep network, a dataset can also be characterized by infinitely many measures. Therefore we do not expect that our choice of the nearest neighbor error rate as a complexity measure is necessarily the best one. As observed in [1], data complexity is multi-faceted in nature; there exist at least several uncorrelated measures each highlighting a different aspect of complexity (e.g. linear separability). In principle, any measure that can well separate the easily learnable problems (e.g. with linearly separable classes) from the knowingly unlearnable problems (e.g. with randomly assigned classes) may serve as a complexity measure – because every classification problem lies somewhere in between. In practice, one may prefer measures that are easily computable or approximated, or well correlated with the accuracies of common classifiers [3].

The data complexity analysis provides a different perspective that can complement other methods for analyzing deep networks, such as studies on their decision boundaries [14], neural tangent kernels and training dynamics [15], inter-layer correlations and representation group flow [16]. The observations obtained here need to be followed up with further studies to compare and cross-check with the observations or predictions from these other approaches.

IX. CONCLUSIONS

We presented an analysis of the representations obtained in a deep neural network using a data complexity measure. We showed that the power of the network in mapping input objects to desirable decision regions is distributed over the network layers. We observed how the complexity of the representations evolves over training epochs, and how it changes due to training with dense or sparse training data. We further showed the effect of using dropout layers to improve robustness of the network. We believe a study like this can help reveal more details about the behavior of such black-box algorithms, and suggest directions for the pursuit of newer approaches to control their behavior. Since our study used well published network designs and datasets, and a very basic complexity measure that does not require complicated tuning, we expect that the results can be easily reproduced, and can serve as a baseline for more advanced studies along these lines.

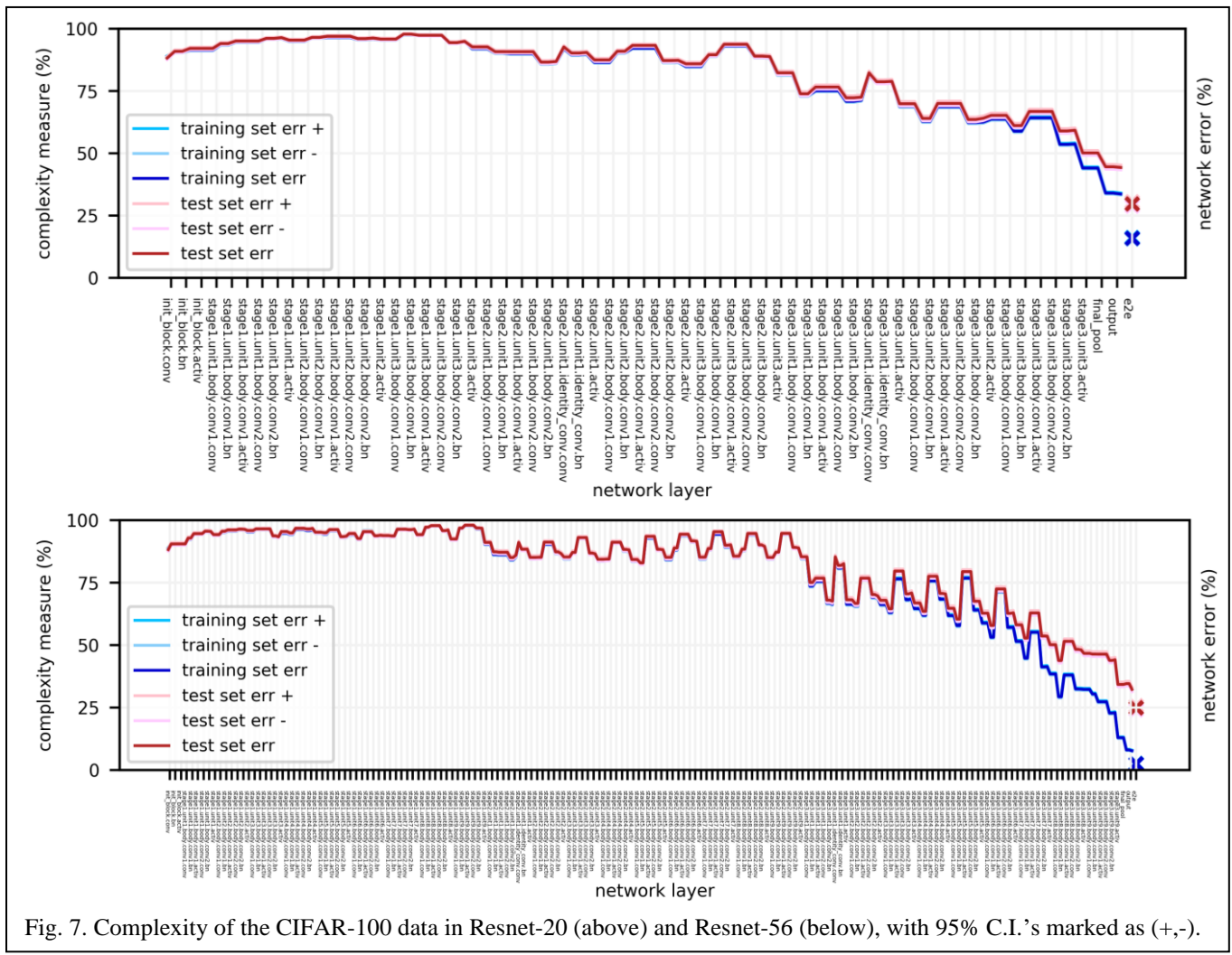

Fig. 7. Complexity of the CIFAR-100 data in Resnet-20 (above) and Resnet-56 (below), with 95% C.I.'s marked as (+,-).

.

# Complexity of Representations in Deep Learning

Tin Kam Ho

**APPENDIX A:**   Architecture of ResNet-20 for CIFAR-100   (resnet20_cifar100)

[Source: O. Sémery, "Computer vision models on PyTorch," https://github.com/osmr/imgclsmob/tree/master/pytorch]

```
Module.parameters of CIFARResNet(
  (features): Sequential(
    (init_block): ConvBlock(
      (conv): Conv2d(3, 16, kernel_size=(3, 3), stride=(1, 1), padding=(1, 1), bias=False)
      (bn): BatchNorm2d(16, eps=1e-05, momentum=0.1, affine=True, track_running_stats=True)
      (activ): ReLU(inplace=True)
    )
    (stage1): Sequential(
      (unit1): ResUnit(
        (body): ResBlock(
          (conv1): ConvBlock(
            (conv): Conv2d(16, 16, kernel_size=(3, 3), stride=(1, 1), padding=(1, 1), bias=False)
            (bn): BatchNorm2d(16, eps=1e-05, momentum=0.1, affine=True, track_running_stats=True)
            (activ): ReLU(inplace=True)
          )
          (conv2): ConvBlock(
            (conv): Conv2d(16, 16, kernel_size=(3, 3), stride=(1, 1), padding=(1, 1), bias=False)
            (bn): BatchNorm2d(16, eps=1e-05, momentum=0.1, affine=True, track_running_stats=True)
          )
        )
        (activ): ReLU(inplace=True)
      )
      (unit2): ResUnit(
        (body): ResBlock(
          (conv1): ConvBlock(
            (conv): Conv2d(16, 16, kernel_size=(3, 3), stride=(1, 1), padding=(1, 1), bias=False)
            (bn): BatchNorm2d(16, eps=1e-05, momentum=0.1, affine=True, track_running_stats=True)
            (activ): ReLU(inplace=True)
          )
          (conv2): ConvBlock(
            (conv): Conv2d(16, 16, kernel_size=(3, 3), stride=(1, 1), padding=(1, 1), bias=False)
            (bn): BatchNorm2d(16, eps=1e-05, momentum=0.1, affine=True, track_running_stats=True)
          )
        )
        (activ): ReLU(inplace=True)
      )
      (unit3): ResUnit(
        (body): ResBlock(
          (conv1): ConvBlock(
            (conv): Conv2d(16, 16, kernel_size=(3, 3), stride=(1, 1), padding=(1, 1), bias=False)
            (bn): BatchNorm2d(16, eps=1e-05, momentum=0.1, affine=True, track_running_stats=True)
            (activ): ReLU(inplace=True)
          )
          (conv2): ConvBlock(
            (conv): Conv2d(16, 16, kernel_size=(3, 3), stride=(1, 1), padding=(1, 1), bias=False)
            (bn): BatchNorm2d(16, eps=1e-05, momentum=0.1, affine=True, track_running_stats=True)
          )
        )
        (activ): ReLU(inplace=True)
      )
    )
    (stage2): Sequential(
      (unit1): ResUnit(
        (body): ResBlock(
          (conv1): ConvBlock(
            (conv): Conv2d(16, 32, kernel_size=(3, 3), stride=(2, 2), padding=(1, 1), bias=False)
            (bn): BatchNorm2d(32, eps=1e-05, momentum=0.1, affine=True, track_running_stats=True)
            (activ): ReLU(inplace=True)
          )
          (conv2): ConvBlock(
```

```
        (conv): Conv2d(32, 32, kernel_size=(3, 3), stride=(1, 1), padding=(1, 1), bias=False)
        (bn): BatchNorm2d(32, eps=1e-05, momentum=0.1, affine=True, track_running_stats=True)
      )
    )
    (identity_conv): ConvBlock(
      (conv): Conv2d(16, 32, kernel_size=(1, 1), stride=(2, 2), bias=False)
      (bn): BatchNorm2d(32, eps=1e-05, momentum=0.1, affine=True, track_running_stats=True)
    )
    (activ): ReLU(inplace=True)
  )
  (unit2): ResUnit(
    (body): ResBlock(
      (conv1): ConvBlock(
        (conv): Conv2d(32, 32, kernel_size=(3, 3), stride=(1, 1), padding=(1, 1), bias=False)
        (bn): BatchNorm2d(32, eps=1e-05, momentum=0.1, affine=True, track_running_stats=True)
        (activ): ReLU(inplace=True)
      )
      (conv2): ConvBlock(
        (conv): Conv2d(32, 32, kernel_size=(3, 3), stride=(1, 1), padding=(1, 1), bias=False)
        (bn): BatchNorm2d(32, eps=1e-05, momentum=0.1, affine=True, track_running_stats=True)
      )
    )
    (activ): ReLU(inplace=True)
  )
  (unit3): ResUnit(
    (body): ResBlock(
      (conv1): ConvBlock(
        (conv): Conv2d(32, 32, kernel_size=(3, 3), stride=(1, 1), padding=(1, 1), bias=False)
        (bn): BatchNorm2d(32, eps=1e-05, momentum=0.1, affine=True, track_running_stats=True)
        (activ): ReLU(inplace=True)
      )
      (conv2): ConvBlock(
        (conv): Conv2d(32, 32, kernel_size=(3, 3), stride=(1, 1), padding=(1, 1), bias=False)
        (bn): BatchNorm2d(32, eps=1e-05, momentum=0.1, affine=True, track_running_stats=True)
      )
    )
    (activ): ReLU(inplace=True)
  )
)
(stage3): Sequential(
  (unit1): ResUnit(
    (body): ResBlock(
      (conv1): ConvBlock(
        (conv): Conv2d(32, 64, kernel_size=(3, 3), stride=(2, 2), padding=(1, 1), bias=False)
        (bn): BatchNorm2d(64, eps=1e-05, momentum=0.1, affine=True, track_running_stats=True)
        (activ): ReLU(inplace=True)
      )
      (conv2): ConvBlock(
        (conv): Conv2d(64, 64, kernel_size=(3, 3), stride=(1, 1), padding=(1, 1), bias=False)
        (bn): BatchNorm2d(64, eps=1e-05, momentum=0.1, affine=True, track_running_stats=True)
      )
    )
    (identity_conv): ConvBlock(
      (conv): Conv2d(32, 64, kernel_size=(1, 1), stride=(2, 2), bias=False)
      (bn): BatchNorm2d(64, eps=1e-05, momentum=0.1, affine=True, track_running_stats=True)
    )
    (activ): ReLU(inplace=True)
  )
  (unit2): ResUnit(
    (body): ResBlock(
      (conv1): ConvBlock(
        (conv): Conv2d(64, 64, kernel_size=(3, 3), stride=(1, 1), padding=(1, 1), bias=False)
        (bn): BatchNorm2d(64, eps=1e-05, momentum=0.1, affine=True, track_running_stats=True)
        (activ): ReLU(inplace=True)
      )
      (conv2): ConvBlock(
        (conv): Conv2d(64, 64, kernel_size=(3, 3), stride=(1, 1), padding=(1, 1), bias=False)
        (bn): BatchNorm2d(64, eps=1e-05, momentum=0.1, affine=True, track_running_stats=True)
      )
    )
    (activ): ReLU(inplace=True)
  )
  (unit3): ResUnit(
    (body): ResBlock(
      (conv1): ConvBlock(
```

```
        (conv): Conv2d(64, 64, kernel_size=(3, 3), stride=(1, 1), padding=(1, 1), bias=False)
        (bn): BatchNorm2d(64, eps=1e-05, momentum=0.1, affine=True, track_running_stats=True)
        (activ): ReLU(inplace=True)
      )
      (conv2): ConvBlock(
        (conv): Conv2d(64, 64, kernel_size=(3, 3), stride=(1, 1), padding=(1, 1), bias=False)
        (bn): BatchNorm2d(64, eps=1e-05, momentum=0.1, affine=True, track_running_stats=True)
      )
    )
    (activ): ReLU(inplace=True)
    )
  )
  (final_pool): AvgPool2d(kernel_size=8, stride=1, padding=0)
)
(output): Linear(in_features=64, out_features=100, bias=True)
)
```

## APPENDIX B:   Architecture of ResNet-56 for CIFAR-100   (resnet56_cifar100)

[Source:  O. Sémery, "Computer vision models on PyTorch," https://github.com/osmr/imgclsmob/tree/master/pytorch]

```
Module.parameters of CIFARResNet(

(features): Sequential(
    (init_block): ConvBlock(
        (conv): Conv2d(3, 16, kernel_size=(3, 3), stride=(1, 1), padding=(1, 1), bias=False)
        (bn): BatchNorm2d(16, eps=1e-05, momentum=0.1, affine=True, track_running_stats=True)
        (activ): ReLU(inplace=True)
    )
    (stage1): Sequential(
      (unit1): ResUnit(
        (body): ResBlock(
          (conv1): ConvBlock(
            (conv): Conv2d(16, 16, kernel_size=(3, 3), stride=(1, 1), padding=(1, 1), bias=False)
            (bn): BatchNorm2d(16, eps=1e-05, momentum=0.1, affine=True, track_running_stats=True)
            (activ): ReLU(inplace=True)
          )
          (conv2): ConvBlock(
            (conv): Conv2d(16, 16, kernel_size=(3, 3), stride=(1, 1), padding=(1, 1), bias=False)
            (bn): BatchNorm2d(16, eps=1e-05, momentum=0.1, affine=True, track_running_stats=True)
          )
        )
        (activ): ReLU(inplace=True)
      )
      (unit2): ResUnit(
        (body): ResBlock(
          (conv1): ConvBlock(
            (conv): Conv2d(16, 16, kernel_size=(3, 3), stride=(1, 1), padding=(1, 1), bias=False)
            (bn): BatchNorm2d(16, eps=1e-05, momentum=0.1, affine=True, track_running_stats=True)
            (activ): ReLU(inplace=True)
          )
          (conv2): ConvBlock(
            (conv): Conv2d(16, 16, kernel_size=(3, 3), stride=(1, 1), padding=(1, 1), bias=False)
            (bn): BatchNorm2d(16, eps=1e-05, momentum=0.1, affine=True, track_running_stats=True)
          )
        )
        (activ): ReLU(inplace=True)
      )
      (unit3): ResUnit(
        (body): ResBlock(
          (conv1): ConvBlock(
            (conv): Conv2d(16, 16, kernel_size=(3, 3), stride=(1, 1), padding=(1, 1), bias=False)
            (bn): BatchNorm2d(16, eps=1e-05, momentum=0.1, affine=True, track_running_stats=True)
            (activ): ReLU(inplace=True)
          )
          (conv2): ConvBlock(
            (conv): Conv2d(16, 16, kernel_size=(3, 3), stride=(1, 1), padding=(1, 1), bias=False)
            (bn): BatchNorm2d(16, eps=1e-05, momentum=0.1, affine=True, track_running_stats=True)
          )
        )
        (activ): ReLU(inplace=True)
```

```
    )
    (unit4): ResUnit(
      (body): ResBlock(
        (conv1): ConvBlock(
          (conv): Conv2d(16, 16, kernel_size=(3, 3), stride=(1, 1), padding=(1, 1), bias=False)
          (bn): BatchNorm2d(16, eps=1e-05, momentum=0.1, affine=True, track_running_stats=True)
          (activ): ReLU(inplace=True)
        )
        (conv2): ConvBlock(
          (conv): Conv2d(16, 16, kernel_size=(3, 3), stride=(1, 1), padding=(1, 1), bias=False)
          (bn): BatchNorm2d(16, eps=1e-05, momentum=0.1, affine=True, track_running_stats=True)
        )
      )
      (activ): ReLU(inplace=True)
    )
    (unit5): ResUnit(
      (body): ResBlock(
        (conv1): ConvBlock(
          (conv): Conv2d(16, 16, kernel_size=(3, 3), stride=(1, 1), padding=(1, 1), bias=False)
          (bn): BatchNorm2d(16, eps=1e-05, momentum=0.1, affine=True, track_running_stats=True)
          (activ): ReLU(inplace=True)
        )
        (conv2): ConvBlock(
          (conv): Conv2d(16, 16, kernel_size=(3, 3), stride=(1, 1), padding=(1, 1), bias=False)
          (bn): BatchNorm2d(16, eps=1e-05, momentum=0.1, affine=True, track_running_stats=True)
        )
      )
      (activ): ReLU(inplace=True)
    )
    (unit6): ResUnit(
      (body): ResBlock(
        (conv1): ConvBlock(
          (conv): Conv2d(16, 16, kernel_size=(3, 3), stride=(1, 1), padding=(1, 1), bias=False)
          (bn): BatchNorm2d(16, eps=1e-05, momentum=0.1, affine=True, track_running_stats=True)
          (activ): ReLU(inplace=True)
        )
        (conv2): ConvBlock(
          (conv): Conv2d(16, 16, kernel_size=(3, 3), stride=(1, 1), padding=(1, 1), bias=False)
          (bn): BatchNorm2d(16, eps=1e-05, momentum=0.1, affine=True, track_running_stats=True)
        )
      )
      (activ): ReLU(inplace=True)
    )
    (unit7): ResUnit(
      (body): ResBlock(
        (conv1): ConvBlock(
          (conv): Conv2d(16, 16, kernel_size=(3, 3), stride=(1, 1), padding=(1, 1), bias=False)
          (bn): BatchNorm2d(16, eps=1e-05, momentum=0.1, affine=True, track_running_stats=True)
          (activ): ReLU(inplace=True)
        )
        (conv2): ConvBlock(
          (conv): Conv2d(16, 16, kernel_size=(3, 3), stride=(1, 1), padding=(1, 1), bias=False)
          (bn): BatchNorm2d(16, eps=1e-05, momentum=0.1, affine=True, track_running_stats=True)
        )
      )
      (activ): ReLU(inplace=True)
    )
    (unit8): ResUnit(
      (body): ResBlock(
        (conv1): ConvBlock(
          (conv): Conv2d(16, 16, kernel_size=(3, 3), stride=(1, 1), padding=(1, 1), bias=False)
          (bn): BatchNorm2d(16, eps=1e-05, momentum=0.1, affine=True, track_running_stats=True)
          (activ): ReLU(inplace=True)
        )
        (conv2): ConvBlock(
          (conv): Conv2d(16, 16, kernel_size=(3, 3), stride=(1, 1), padding=(1, 1), bias=False)
          (bn): BatchNorm2d(16, eps=1e-05, momentum=0.1, affine=True, track_running_stats=True)
        )
      )
      (activ): ReLU(inplace=True)
    )
    (unit9): ResUnit(
      (body): ResBlock(
        (conv1): ConvBlock(
          (conv): Conv2d(16, 16, kernel_size=(3, 3), stride=(1, 1), padding=(1, 1), bias=False)
```

```
          (bn): BatchNorm2d(16, eps=1e-05, momentum=0.1, affine=True, track_running_stats=True)
          (activ): ReLU(inplace=True)
        )
        (conv2): ConvBlock(
          (conv): Conv2d(16, 16, kernel_size=(3, 3), stride=(1, 1), padding=(1, 1), bias=False)
          (bn): BatchNorm2d(16, eps=1e-05, momentum=0.1, affine=True, track_running_stats=True)
        )
      )
      (activ): ReLU(inplace=True)
    )
  )
  (stage2): Sequential(
    (unit1): ResUnit(
      (body): ResBlock(
        (conv1): ConvBlock(
          (conv): Conv2d(16, 32, kernel_size=(3, 3), stride=(2, 2), padding=(1, 1), bias=False)
          (bn): BatchNorm2d(32, eps=1e-05, momentum=0.1, affine=True, track_running_stats=True)
          (activ): ReLU(inplace=True)
        )
        (conv2): ConvBlock(
          (conv): Conv2d(32, 32, kernel_size=(3, 3), stride=(1, 1), padding=(1, 1), bias=False)
          (bn): BatchNorm2d(32, eps=1e-05, momentum=0.1, affine=True, track_running_stats=True)
        )
      )
      (identity_conv): ConvBlock(
        (conv): Conv2d(16, 32, kernel_size=(1, 1), stride=(2, 2), bias=False)
        (bn): BatchNorm2d(32, eps=1e-05, momentum=0.1, affine=True, track_running_stats=True)
      )
      (activ): ReLU(inplace=True)
    )
    (unit2): ResUnit(
      (body): ResBlock(
        (conv1): ConvBlock(
          (conv): Conv2d(32, 32, kernel_size=(3, 3), stride=(1, 1), padding=(1, 1), bias=False)
          (bn): BatchNorm2d(32, eps=1e-05, momentum=0.1, affine=True, track_running_stats=True)
          (activ): ReLU(inplace=True)
        )
        (conv2): ConvBlock(
          (conv): Conv2d(32, 32, kernel_size=(3, 3), stride=(1, 1), padding=(1, 1), bias=False)
          (bn): BatchNorm2d(32, eps=1e-05, momentum=0.1, affine=True, track_running_stats=True)
        )
      )
      (activ): ReLU(inplace=True)
    )
    (unit3): ResUnit(
      (body): ResBlock(
        (conv1): ConvBlock(
          (conv): Conv2d(32, 32, kernel_size=(3, 3), stride=(1, 1), padding=(1, 1), bias=False)
          (bn): BatchNorm2d(32, eps=1e-05, momentum=0.1, affine=True, track_running_stats=True)
          (activ): ReLU(inplace=True)
        )
        (conv2): ConvBlock(
          (conv): Conv2d(32, 32, kernel_size=(3, 3), stride=(1, 1), padding=(1, 1), bias=False)
          (bn): BatchNorm2d(32, eps=1e-05, momentum=0.1, affine=True, track_running_stats=True)
        )
      )
      (activ): ReLU(inplace=True)
    )
    (unit4): ResUnit(
      (body): ResBlock(
        (conv1): ConvBlock(
          (conv): Conv2d(32, 32, kernel_size=(3, 3), stride=(1, 1), padding=(1, 1), bias=False)
          (bn): BatchNorm2d(32, eps=1e-05, momentum=0.1, affine=True, track_running_stats=True)
          (activ): ReLU(inplace=True)
        )
        (conv2): ConvBlock(
          (conv): Conv2d(32, 32, kernel_size=(3, 3), stride=(1, 1), padding=(1, 1), bias=False)
          (bn): BatchNorm2d(32, eps=1e-05, momentum=0.1, affine=True, track_running_stats=True)
        )
      )
      (activ): ReLU(inplace=True)
    )
    (unit5): ResUnit(
      (body): ResBlock(
        (conv1): ConvBlock(
```

```
        (conv): Conv2d(32, 32, kernel_size=(3, 3), stride=(1, 1), padding=(1, 1), bias=False)
        (bn): BatchNorm2d(32, eps=1e-05, momentum=0.1, affine=True, track_running_stats=True)
        (activ): ReLU(inplace=True)
      )
      (conv2): ConvBlock(
        (conv): Conv2d(32, 32, kernel_size=(3, 3), stride=(1, 1), padding=(1, 1), bias=False)
        (bn): BatchNorm2d(32, eps=1e-05, momentum=0.1, affine=True, track_running_stats=True)
      )
    )
    (activ): ReLU(inplace=True)
  )
  (unit6): ResUnit(
    (body): ResBlock(
      (conv1): ConvBlock(
        (conv): Conv2d(32, 32, kernel_size=(3, 3), stride=(1, 1), padding=(1, 1), bias=False)
        (bn): BatchNorm2d(32, eps=1e-05, momentum=0.1, affine=True, track_running_stats=True)
        (activ): ReLU(inplace=True)
      )
      (conv2): ConvBlock(
        (conv): Conv2d(32, 32, kernel_size=(3, 3), stride=(1, 1), padding=(1, 1), bias=False)
        (bn): BatchNorm2d(32, eps=1e-05, momentum=0.1, affine=True, track_running_stats=True)
      )
    )
    (activ): ReLU(inplace=True)
  )
  (unit7): ResUnit(
    (body): ResBlock(
      (conv1): ConvBlock(
        (conv): Conv2d(32, 32, kernel_size=(3, 3), stride=(1, 1), padding=(1, 1), bias=False)
        (bn): BatchNorm2d(32, eps=1e-05, momentum=0.1, affine=True, track_running_stats=True)
        (activ): ReLU(inplace=True)
      )
      (conv2): ConvBlock(
        (conv): Conv2d(32, 32, kernel_size=(3, 3), stride=(1, 1), padding=(1, 1), bias=False)
        (bn): BatchNorm2d(32, eps=1e-05, momentum=0.1, affine=True, track_running_stats=True)
      )
    )
    (activ): ReLU(inplace=True)
  )
  (unit8): ResUnit(
    (body): ResBlock(
      (conv1): ConvBlock(
        (conv): Conv2d(32, 32, kernel_size=(3, 3), stride=(1, 1), padding=(1, 1), bias=False)
        (bn): BatchNorm2d(32, eps=1e-05, momentum=0.1, affine=True, track_running_stats=True)
        (activ): ReLU(inplace=True)
      )
      (conv2): ConvBlock(
        (conv): Conv2d(32, 32, kernel_size=(3, 3), stride=(1, 1), padding=(1, 1), bias=False)
        (bn): BatchNorm2d(32, eps=1e-05, momentum=0.1, affine=True, track_running_stats=True)
      )
    )
    (activ): ReLU(inplace=True)
  )
  (unit9): ResUnit(
    (body): ResBlock(
      (conv1): ConvBlock(
        (conv): Conv2d(32, 32, kernel_size=(3, 3), stride=(1, 1), padding=(1, 1), bias=False)
        (bn): BatchNorm2d(32, eps=1e-05, momentum=0.1, affine=True, track_running_stats=True)
        (activ): ReLU(inplace=True)
      )
      (conv2): ConvBlock(
        (conv): Conv2d(32, 32, kernel_size=(3, 3), stride=(1, 1), padding=(1, 1), bias=False)
        (bn): BatchNorm2d(32, eps=1e-05, momentum=0.1, affine=True, track_running_stats=True)
      )
    )
    (activ): ReLU(inplace=True)
  )
)
(stage3): Sequential(
  (unit1): ResUnit(
    (body): ResBlock(
      (conv1): ConvBlock(
        (conv): Conv2d(32, 64, kernel_size=(3, 3), stride=(2, 2), padding=(1, 1), bias=False)
        (bn): BatchNorm2d(64, eps=1e-05, momentum=0.1, affine=True, track_running_stats=True)
        (activ): ReLU(inplace=True)
```

```
      )
      (conv2): ConvBlock(
        (conv): Conv2d(64, 64, kernel_size=(3, 3), stride=(1, 1), padding=(1, 1), bias=False)
        (bn): BatchNorm2d(64, eps=1e-05, momentum=0.1, affine=True, track_running_stats=True)
      )
    )
    (identity_conv): ConvBlock(
      (conv): Conv2d(32, 64, kernel_size=(1, 1), stride=(2, 2), bias=False)
      (bn): BatchNorm2d(64, eps=1e-05, momentum=0.1, affine=True, track_running_stats=True)
    )
    (activ): ReLU(inplace=True)
  )
  (unit2): ResUnit(
    (body): ResBlock(
      (conv1): ConvBlock(
        (conv): Conv2d(64, 64, kernel_size=(3, 3), stride=(1, 1), padding=(1, 1), bias=False)
        (bn): BatchNorm2d(64, eps=1e-05, momentum=0.1, affine=True, track_running_stats=True)
        (activ): ReLU(inplace=True)
      )
      (conv2): ConvBlock(
        (conv): Conv2d(64, 64, kernel_size=(3, 3), stride=(1, 1), padding=(1, 1), bias=False)
        (bn): BatchNorm2d(64, eps=1e-05, momentum=0.1, affine=True, track_running_stats=True)
      )
    )
    (activ): ReLU(inplace=True)
  )
  (unit3): ResUnit(
    (body): ResBlock(
      (conv1): ConvBlock(
        (conv): Conv2d(64, 64, kernel_size=(3, 3), stride=(1, 1), padding=(1, 1), bias=False)
        (bn): BatchNorm2d(64, eps=1e-05, momentum=0.1, affine=True, track_running_stats=True)
        (activ): ReLU(inplace=True)
      )
      (conv2): ConvBlock(
        (conv): Conv2d(64, 64, kernel_size=(3, 3), stride=(1, 1), padding=(1, 1), bias=False)
        (bn): BatchNorm2d(64, eps=1e-05, momentum=0.1, affine=True, track_running_stats=True)
      )
    )
    (activ): ReLU(inplace=True)
  )
  (unit4): ResUnit(
    (body): ResBlock(
      (conv1): ConvBlock(
        (conv): Conv2d(64, 64, kernel_size=(3, 3), stride=(1, 1), padding=(1, 1), bias=False)
        (bn): BatchNorm2d(64, eps=1e-05, momentum=0.1, affine=True, track_running_stats=True)
        (activ): ReLU(inplace=True)
      )
      (conv2): ConvBlock(
        (conv): Conv2d(64, 64, kernel_size=(3, 3), stride=(1, 1), padding=(1, 1), bias=False)
        (bn): BatchNorm2d(64, eps=1e-05, momentum=0.1, affine=True, track_running_stats=True)
      )
    )
    (activ): ReLU(inplace=True)
  )
  (unit5): ResUnit(
    (body): ResBlock(
      (conv1): ConvBlock(
        (conv): Conv2d(64, 64, kernel_size=(3, 3), stride=(1, 1), padding=(1, 1), bias=False)
        (bn): BatchNorm2d(64, eps=1e-05, momentum=0.1, affine=True, track_running_stats=True)
        (activ): ReLU(inplace=True)
      )
      (conv2): ConvBlock(
        (conv): Conv2d(64, 64, kernel_size=(3, 3), stride=(1, 1), padding=(1, 1), bias=False)
        (bn): BatchNorm2d(64, eps=1e-05, momentum=0.1, affine=True, track_running_stats=True)
      )
    )
    (activ): ReLU(inplace=True)
  )
  (unit6): ResUnit(
    (body): ResBlock(
      (conv1): ConvBlock(
        (conv): Conv2d(64, 64, kernel_size=(3, 3), stride=(1, 1), padding=(1, 1), bias=False)
        (bn): BatchNorm2d(64, eps=1e-05, momentum=0.1, affine=True, track_running_stats=True)
        (activ): ReLU(inplace=True)
      )
```

```
        (conv2): ConvBlock(
          (conv): Conv2d(64, 64, kernel_size=(3, 3), stride=(1, 1), padding=(1, 1), bias=False)
          (bn): BatchNorm2d(64, eps=1e-05, momentum=0.1, affine=True, track_running_stats=True)
        )
      )
      (activ): ReLU(inplace=True)
    )
    (unit7): ResUnit(
      (body): ResBlock(
        (conv1): ConvBlock(
          (conv): Conv2d(64, 64, kernel_size=(3, 3), stride=(1, 1), padding=(1, 1), bias=False)
          (bn): BatchNorm2d(64, eps=1e-05, momentum=0.1, affine=True, track_running_stats=True)
          (activ): ReLU(inplace=True)
        )
        (conv2): ConvBlock(
          (conv): Conv2d(64, 64, kernel_size=(3, 3), stride=(1, 1), padding=(1, 1), bias=False)
          (bn): BatchNorm2d(64, eps=1e-05, momentum=0.1, affine=True, track_running_stats=True)
        )
      )
      (activ): ReLU(inplace=True)
    )
    (unit8): ResUnit(
      (body): ResBlock(
        (conv1): ConvBlock(
          (conv): Conv2d(64, 64, kernel_size=(3, 3), stride=(1, 1), padding=(1, 1), bias=False)
          (bn): BatchNorm2d(64, eps=1e-05, momentum=0.1, affine=True, track_running_stats=True)
          (activ): ReLU(inplace=True)
        )
        (conv2): ConvBlock(
          (conv): Conv2d(64, 64, kernel_size=(3, 3), stride=(1, 1), padding=(1, 1), bias=False)
          (bn): BatchNorm2d(64, eps=1e-05, momentum=0.1, affine=True, track_running_stats=True)
        )
      )
      (activ): ReLU(inplace=True)
    )
    (unit9): ResUnit(
      (body): ResBlock(
        (conv1): ConvBlock(
          (conv): Conv2d(64, 64, kernel_size=(3, 3), stride=(1, 1), padding=(1, 1), bias=False)
          (bn): BatchNorm2d(64, eps=1e-05, momentum=0.1, affine=True, track_running_stats=True)
          (activ): ReLU(inplace=True)
        )
        (conv2): ConvBlock(
          (conv): Conv2d(64, 64, kernel_size=(3, 3), stride=(1, 1), padding=(1, 1), bias=False)
          (bn): BatchNorm2d(64, eps=1e-05, momentum=0.1, affine=True, track_running_stats=True)
        )
      )
      (activ): ReLU(inplace=True)
    )
  )
  (final_pool): AvgPool2d(kernel_size=8, stride=1, padding=0)
  )
  (output): Linear(in_features=64, out_features=100, bias=True)
)
```